\pdfoutput=1

\documentclass[11pt]{article}

\usepackage{ACL2023}

\usepackage{times}
\usepackage{latexsym}

\usepackage[T1]{fontenc}

\usepackage[utf8]{inputenc}

\usepackage{microtype}

\usepackage{inconsolata}
\usepackage{rotating}
\usepackage{makecell}
\usepackage{diagbox}

\title{A Call for Standardization and Validation of\\Text Style Transfer Evaluation}

\author{Phil Ostheimer, Mayank Nagda, Marius Kloft, Sophie Fellenz\\
    surname@cs.uni-kl.de \\ RPTU Kaiserslautern-Landau}

\begin{document}
\maketitle
\begin{abstract}
Text Style Transfer (TST) evaluation is, in practice, inconsistent.
Therefore, we conduct a meta-analysis on human and automated TST evaluation and experimentation that thoroughly examines existing literature in the field.
The meta-analysis reveals a substantial standardization gap in human and automated evaluation.
In addition, we also find a validation gap: only few automated metrics have been validated using human experiments.
To this end, we thoroughly scrutinize both the standardization and validation gap and reveal the resulting pitfalls.
This work also paves the way to close the standardization and validation gap in TST evaluation by calling out requirements to be met by future research.
\end{abstract}

\section{Introduction}
Text style transfer (TST) is the task of transferring text from one style to another. Examples include sentiment transfer (making a negative text more positive) \citep{Shen:2017}, politeness transfer \citep{Niu:2018}, formality transfer \citep{Rao:2018}, and many more. Anyone working and communicating with texts has most likely manually performed many of these transfer tasks countless times. Whether generalized \citep{Reif:2022} or task-specific models, evaluating their performance on TST is crucial to measure progress in task-specific text generation. In the last few years, there has been a surge in research on TST (see Figure~\ref{fig:publications}, Appendix), the majority of which was in the last two years. When a field develops that quickly, the development of evaluation procedures often does not keep up.

TST research is now at a point where reflection on evaluation practices is becoming increasingly urgent, as progress in this field is becoming difficult if the performance of new methods cannot be reliably compared to existing methods. It can even become hard to determine the state of the art one should aim to improve. However, recent TST surveys \cite{Toshevska:2022, Jin:2022, Hu:2022} focus on summarizing TST methods and their evaluation but do not point out evaluation and experimentation issues. Works on TST evaluation either focus on getting more reliable metrics for a particular type of TST \cite{Mir:2019}, focus on a specific evaluation aspect for validation \cite{Yamshchikov:2021}, or emphasize only shortcomings of automated evaluation in a multilingual setting of one particular type of TST \cite{Briakou:2021b}. As a first step, it is, therefore, necessary to examine evaluation practices and experimental setups to point out the need for standardization of both, which is what we set out to do in this paper.

Early TST works \citep{Shen:2017, Li:2018} focus on human evaluation, which is still considered the most reliable \cite{Briakou:2021}. Many later publications rely only on automated metrics, which can be validated by correlating them with human evaluations. So far, no comprehensive overview exists showing which metrics are validated and how leading to researchers using popular metrics rather than those with the highest correlations. In our meta-analysis, we counter-act this by reporting which metrics are validated and analyzing cases where the same metric received different validation results. To sum up, our contributions are:

\begin{enumerate}
    \item Our examination of TST methods and TST evaluation practices highlights the need for standardization (Section~\ref{sec:standardization_gap}).
    \item Our overview of automated TST evaluation reveals an unmet need for validation (Section~\ref{sec:validation_gap}).
\end{enumerate}

We conclude with a list of requirements to help close these two gaps.

\section{A Meta-Analysis of Text Style Transfer}
\label{sec:standardization_gap}

In this meta-analysis, we focus on TST publications in top-tier NLP and AI venues (see Appendix \ref{sec:paper_selection} for selection details), resulting in 89 considered papers summarized in Table~ \ref{tab:meta_analysis_1}, \ref{tab:meta_analysis_2}, and \ref{tab:meta_analysis_3}. The dimensions of TST evaluation are fluency, content preservation, and style transfer strength, as well as other aspects. Therefore, we divide the automated and human evaluation subsections according to these aspects. 

\subsection{Automated Evaluation}
\label{sec:automated_evaluation}
For automated evaluation, we aggregate the findings of our meta-analysis in Table~\ref{tab:automated_eval} (with a detailed description of conventions in Appendix~\ref{sec:automated_eval_conventions}). Overall, 21/89 papers use only automated evaluation, and 33/89 use at least one metric that has not been validated. Statistically significant results are reported for only 3/89 papers.

\begin{table}[th]
    \centering
        \begin{tabular}{p{0.35\textwidth}r}
            \textbf{Aspect} & \textbf{Count} \\
            \hline
            No human evaluation & 21/89\\
            >1 non-validated metric & 33/89\\
            Statistical significance & 3/89\\
            Metrics & \\
            \hspace{5mm}Fluency & 34 \\
            \hspace{5mm}Content preservation & 35 \\
            \hspace{5mm}Style transfer strength & 23\\
            \hspace{5mm}Other & 24\\
            Papers & \\
            \hspace{5mm}Fluency & 45/89\\
            \hspace{5mm}Content preservation & 66/89\\
            \hspace{5mm}Style transfer strength & 77/89\\
            \hspace{5mm}Other & 34/89\\
            \hline
        \end{tabular}
    \caption{Shown is the variety of automated TST evaluation metrics. All examined TST papers deploy at least one. However, using non-validated metrics is a common phenomenon, and statistical significance is reported only in a fraction of the investigated papers.}
    \label{tab:automated_eval}
\end{table}

\paragraph{Fluency} \label{sec:fluency} We find a total of 34 different setups to measure fluency automatically. 45/89 papers rely on automated metrics for measuring fluency. The by far most common metric (see Table~\ref{tab:automated_fluency_ranking}) is the Perplexity (PPL) of a Language Model (LM). However, the pre-training procedure of the LM varies. It is either trained on all styles of a dataset \cite{Santos:2018, Dai:2019, John:2019, Cheng:2020}, trained on each style separately \cite{Yang:2018, Gong:2019}, pre-trained on external datasets \cite{Logeswaran:2018, Jain:2019}, or trained on the dataset and the TST output (backward/forward) \cite{Zhao:2018, Huang:2020}. 19/89 papers do not mention how the LM is trained. These numbers show the need for consistency.

\paragraph{Content Preservation} We find 66 papers using a total of 35 different automated metrics to measure content preservation. Only 15 of these metrics have been validated. 

The most common metrics (see Table~\ref{tab:automated_content_ranking}) to measure content preservation are Source-BLEU and Ref-BLEU (applying BLEU \cite{Papineni:2002} to the input/output, reference/output respectively). However, as pointed out by \citet{Mir:2019}, Source-BLEU only measures n-gram overlaps and does not consider that a change of the sentence is necessary to change the style. This results in contradictory validation results (see Section~\ref{sec:validation_gap}). \citet{Post:2018} has shown that the reported BLEU scores heavily depend on preprocessing and several parameters (e.g., number of references, length penalty, maximum n-gram length, smoothing applied to 0-count n-grams). Works using BLEU for evaluation need this specification to make the results consistent and reproducible. In our meta-analysis, we could not find a single TST paper that specifies BLEU sufficiently (including above mentioned details). Ref-BLEU is the second most popular method to measure content preservation. In addition to the general problems with BLEU, we see two more shortcomings. On the one hand, the needed reference sentences are not available for all datasets. On the other hand, calculating BLEU scores between the output and multiple human references can improve its reliability and, thankfully, for the arguably most popular TST dataset---Yelp---\citet{Jin:2019} and \citet{Luo:2019} introduced additional reference sentences. These, however, are only used by 6/30 papers applying Ref-BLEU on Yelp.

\paragraph{Style Transfer Accuracy} For automated evaluation of style transfer accuracy, we find 23 evaluation metrics, of which 14 have been validated. Overall, 77 papers use automated style transfer accuracy metrics. Table~\ref{tab:automated_style_ranking} provides an overview of the most common automated text style transfer strength metrics. The top two are TextCNN \cite{Kim:2014} and fastText \cite{Joulin:2017}. 

\paragraph{Other} 
34/89 papers measure a fourth aspect. For 29/89 it is an overall metric, and 8/89 measure another aspect of TST. We find eight validated metrics to measure overall performance. None of the metrics for other aspects have been validated.

\subsection{Human Evaluation}
\begin{table}[ht!]
    \centering
        \begin{tabular}{p{0.35\textwidth}r}
            \textbf{Aspect} & \textbf{Count} \\
            \hline
            Usage & 68/89\\
            Statistical analysis & 2/68\\
            Evaluations released & 5/68\\
            Underspecified & 67/68\\
            No. of evaluation schemes & 24\\
            \hline
        \end{tabular}
    \caption{Shown are the aggregated insights of human TST evaluation. Despite being widespread, it is far from being standardized and, in most cases, lacks statistical analysis. It has many different setups that are often underspecified, and the annotations are not released.}
    \label{tab:human_eval}
\end{table}
For human evaluation, we aggregate the findings of our meta-analysis in Table~\ref{tab:human_eval} (with a detailed description of our conventions in Appendix~\ref{sec:human_evaluation_conventions}). Overall, 68/89 papers conduct a human evaluation. However, only a tiny fraction of 2/68 papers detect a statistically significant difference in model performance. Most papers fail to include a statistical test, be it a statistical power analysis before conducting the human evaluations or a statistical significance test afterward. This is a common problem in the NLP community \cite{Card:2020}.

Releasing human evaluations is also relatively uncommon. Only 5/68 papers publish their human experimentation results. However, releasing them would facilitate reproducibility and the validation of new automated metrics. Reproducibility of human TST evaluation is a challenge, as is reproducibility of human evaluation in general \cite{Belz:2020}. 67/68 papers conducting human evaluations have no adequate description. We consider the description of human evaluation to be adequate if the following is specified: annotators' background, the number of annotators, the number of annotators per sample, incentivization, data selection, questionnaire design, and rating scale \cite{Briakou:2021}. With published human evaluations, one could also easily estimate the parameters of a simulation-based power analysis as done by \citet{Card:2020}.

Overall, there is still room for improvement despite previous calls for standardization \cite{Briakou:2021}.
For human evaluation, we find a total of 24 different evaluation schemes (viewed on a high level whether each aspect is present and evaluated relatively or absolutely, not considering different rating scales and questionnaire designs).

\subsection{Experimentation}
\begin{table}[ht!]
    \centering
        \begin{tabular}{p{0.35\textwidth}r}
            \textbf{Aspect} & \textbf{Count} \\
            \hline
            Multiple runs & 5/89\\
            Reproducibility & \\
            \hspace{5mm}Code provided & 56/89\\
            \hspace{5mm}Evaluation code provided & 42/89\\
            Preprocessing specified & 38/89\\
            \hline
        \end{tabular}
    \caption{Shown are the aggregated insights of TST experimentation: a lack of reporting multiple runs, hampered reproducibility by the missing provision of code, and the underspecification of the preprocessing pipeline.}
    \label{tab:experimentation}
\end{table}

We aggregate our meta-analysis' findings on experimentation details in Table~\ref{tab:experimentation} (with a detailed description of our conventions in Appendix~\ref{sec:experimentation_conventions}). In order to make statements about relative model performance, one usually runs the model multiple times with different seeds to be able to conduct a statistical significance test. A TST model's performance can significantly vary between runs \cite{Tikhonov:2019, Yu:2021}, indicating the need for reporting multiple runs. However, most (84/89) papers only report a single run.

Reproducing results is difficult since only 56/89 of the reviewed papers release their code. An even smaller fraction provides the complete evaluation code (42/89). Another aspect that can significantly influence the outcome is preprocessing. However, only 38/89 papers specify their preprocessing pipeline.

\section{Automated Metrics and Their (Missing) Validation}
\label{sec:validation_gap}
In this section, we summarize existing automated metrics and their validity in terms of correlation with human evaluations. 
In Table~\ref{tab:val_overview}, we give a detailed overview (the first of its kind) and describe our conventions in Appendix~\ref{sec:human_validation_details}. The most crucial convention is to assume that the validation of an automated metric for a particular TST task also generalizes to other TST tasks.

\subsection{Fluency}
\label{sec:automated_fluency}
Fluency is sometimes referred to as grammaticality, readability, or naturalness \cite{Mir:2019}. It is commonly quantified by measuring the PPL of an LM on the TST output (40/45 reviewed papers). \citet{Mir:2019} claim a limited correlation between sentence-level Long Short-Term Memory (LSTM) \cite{Hochreiter:1997} LM PPL and human fluency evaluations and conclude that LM PPL is an inappropriate metric to measure fluency in the TST setting. On the other hand, \citet{Pang:2019} show a high correlation but also note that PPL is not equal to fluency. \citet{Briakou:2021b} report a relatively low correlation for the PPL of a 5-gram KenLM \cite{Heafield:2011} with human evaluations and slightly higher correlations for Pseudo Log-Likelihoods (PLL) of BERT \cite{Devlin:2019}, and XLM \cite{Conneau:2020}. Overall, LM PPL for TST evaluation has only been validated on a fraction of the deployed TST datasets, namely Yelp[s] and GYAFC. Previous work is divided as to whether and to what extent this metric correlates with human evaluations. As reported in Section~\ref{sec:standardization_gap}, many different architectures, training methods, and application methods of LMs for TST fluency evaluation exist. However, as a simplification, we assume the LM PPL to be a validated fluency metric.

Other approaches to evaluate fluency are based on BLEU. In contrast to \citet{Luo:2019}, \citet{Li:2018} show no significant correlation for Ref-BLEU. \citet{Pryzant:2020} report a low correlation of the Source-BLEU score with human evaluations. \citet{Wu:2020} and \citet{Rao:2018} report moderate correlation for their metric \cite{Heilmann:2014}. \citet{Pryzant:2020} show that their style classifier correlates more with human evaluations of fluency than the style transfer strength. None of these metrics is deployed by more than two papers.
 
\subsection{Content Preservation}
There are two extensive studies for automated content preservation metrics by \citet{Mir:2019} and \citet{Yamshchikov:2021}. However, both have limited scope:  \citet{Mir:2019} only report scores where style words have been masked or removed (not done by any other paper). \citet{Yamshchikov:2021} report correlations only on the datasets themselves and not on actual model outputs. Both report Word Mover's Distance (WMD) \cite{Kusner:2015} having the highest correlation, outperforming Source-BLEU, other embedding-based metrics (such as also investigated by \citet{Fu:2018} and \citet{Pang:2019}), and chrF \cite{Popovic:2015}. 

\citet{Cao:2020} find high correlation for Source-BLEU, whereas \citet{Pryzant:2020} find low correlation. For Ref-BLEU, \citet{Li:2018}, \citet{Luo:2019}, and \citet{Cao:2020} show a high, and \citet{Xu:2012} a low correlation, whereas \citet{Briakou:2021b} investigate Ref-BLEU and Source-BLEU among others, but show chrF having the highest correlation. The suitability of BLEU for TST evaluation remains questionable as \citet{Mir:2019} point out that BLEU cannot capture whether words were changed on purpose.

\subsection{Style Transfer Strength}
Style transfer strength is usually evaluated by applying a sentence classifier trained to classify the output sentences by style. Early work \cite{Xu:2012} compares several metrics showing the highest correlation for Logistic Regression (LR). The two most popular methods nowadays are TextCNN \cite{Kim:2014} and fastText \cite{Joulin:2017}. \citet{Luo:2019} show high correlation for TextCNN, \citet{Pang:2019} for fastText, whereas \citet{Mir:2019} validate both, showing slightly better correlations for TextCNN. \citet{Li:2018} validate a Bi-directional LSTM for style classification and note that the correlation to human evaluations highly depends on the dataset. \citet{Rao:2018} report a moderate correlation to their style transfer strength metric.
Also, \citet{Mir:2019} show that the Earth Mover's Distance (EMD) in combination with TextCNN or fastText has a higher correlation with human evaluations than only TextCNN or fastText, but no other paper uses it.

\subsection{Other}
\citet{Niu:2018} report a high correlation of Source-BLEU with overall dialogue quality. \citet{Rao:2018} report a relatively high correlation for Ref-BLEU compared to TERp and PINC with the overall human evaluations. This is in agreement with \citet{Wang:2020}. \citet{Wu:2020} perform dataset-dependent studies and found no significant correlation between automated metrics and human scores.

\subsection{Metrics Validated for Multiple Aspects}
Some metrics, such as Ref-BLEU, Source-BLEU, PINC, and embedding-based metrics, are validated for multiple aspects. However, Ref-BLEU shows the highest correlation as an overall metric only for \citet{Rao:2018} (also outperforming PINC), otherwise \cite{Xu:2012, Li:2018, Luo:2019, Cao:2020, Wang:2020, Briakou:2021b}, there is no clear picture. For Source-BLEU \cite{Niu:2018, Mir:2019, Pryzant:2020, Cao:2020, Yamshchikov:2021, Briakou:2021b} and also for embedding-based metrics \cite{Xu:2012,Fu:2018,Mir:2019,Pang:2019,Wu:2020,Yamshchikov:2021,Briakou:2021b}, we find mixed results across the different aspects.

\section{Conclusion \& Future Work}
Our research emphasizes the pressing need for standardization and validation in TST. While human evaluation is still considered to be the most reliable, it is expensive and time-consuming, hindering reproducibility. Many publications use automated metrics as surrogates, but only a fraction of them is validated. Furthermore, human and automated evaluations and experimental setups lack standardization, making it difficult to compare the performance of different models. Summarizing our results, we pose the following six requirements to be met by future TST research: 
\begin{enumerate}
    \item It needs to use validated metrics (see Table~\ref{tab:val_overview}), focusing on those showing the highest correlations with human evaluations
    \item In experiments, multiple runs need to be performed on different random seeds, reporting mean and standard deviation.
    \item A statistical significance test needs to be performed on the results of automated metrics. 
    \item If a human evaluation is done, a statistical power analysis is necessary in advance, and all human evaluation details need to be published as suggested by \citet{Briakou:2021}.
    \item To improve reproducibility, researchers should always specify the preprocessing pipeline and publish their code (including evaluation code).
    \item A comparison with state-of-the-art methods on the validated metrics is called for.
\end{enumerate}
To help with these requirements, we plan a large-scale experimental comparison to rank existing methods according to validated metrics in the future and examine how automated metrics generalize from one TST task to the other. We also plan to (re-)validate existing automated metrics to help meet the first requirement.

\section*{Limitations}
The present work only points out problems of existing research and presents no final solutions. 
We also simplify the assumption that an automated metric validated for a specific TST task generalizes to other tasks. However, this is problematic since there is, to our knowledge, no investigation of whether a validation on one task generalizes. This concern is motivated by the fundamental differences in how different TST tasks are defined. There are several different definitions, such as data-driven TST (e.g., sentiment transfer) and linguistically motivated TST (e.g., formality transfer) \cite{Jin:2022}. Also, we consider only TST papers (no text simplification) and focus on top-tier NLP and AI venues (non-workshop).

\section*{Acknowledgements}
The authors acknowledge support by the Carl-Zeiss Foundation, the BMBF awards 01|S20048, 03|B0770E, and 01|S21010C, and the DFG awards BU 4042/2-1, BU 4042/1-1, KL 2698/2-1, KL 2698/5-1, KL 2698/6-1, and KL 2698/7-1.

\bibliography{custom}
\bibliographystyle{acl_natbib}

\appendix

\section{Meta-Analysis}
\label{sec:appendix}
This section describes the details of the conducted meta-analysis on TST evaluation.

\subsection{Popularity of TST}
\begin{figure}[th]
\centering
\includegraphics[width=0.9\columnwidth]{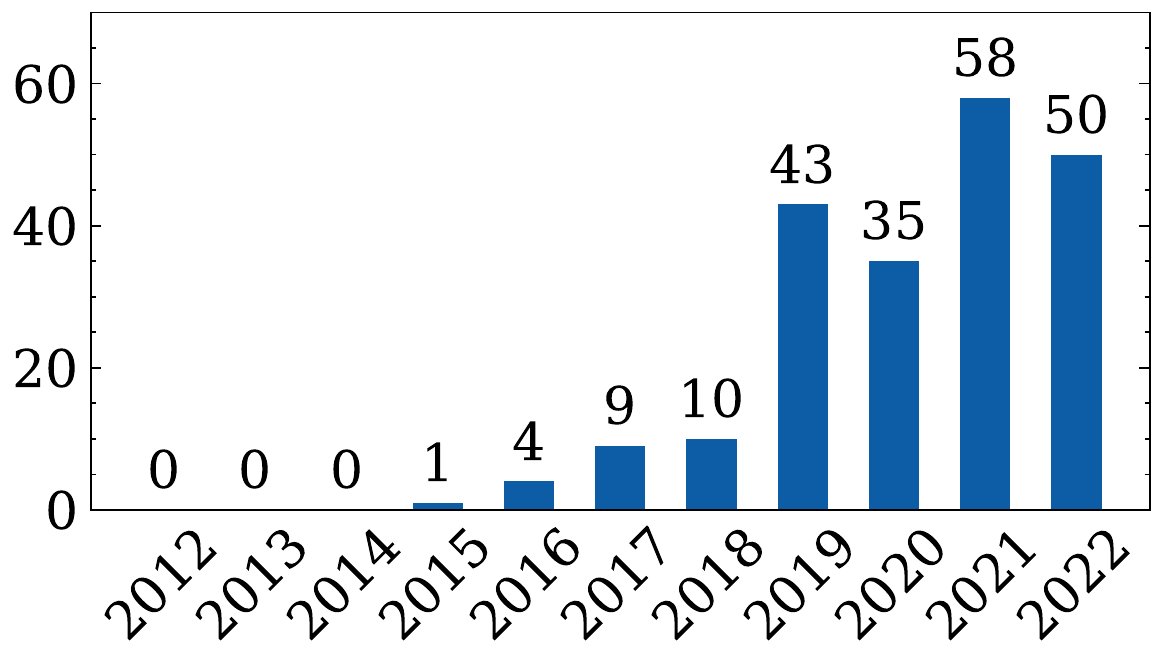} 
\caption[The number of TST publications has steeply increased in the last ten years]{The number of TST publications has steeply increased in the last ten years.}
\label{fig:publications}
\end{figure}

TST has become a popular topic in research. When Searching for "text style transfer" on \url{dblp.uni-trier.de} (accessed Jan 18, 2023), we can find a steep increase in the number of publications per year (see Figure~\ref{fig:publications}).

\subsection{Paper Selection}
\label{sec:paper_selection}
We consider the papers listed in a recent TST survey \cite{Jin:2022} \footnote{\url{https://github.com/fuzhenxin/Style-Transfer-in-Text}} with the addition of more recent publications. We consider only TST papers (no text simplification) and focus on top-tier NLP and AI venues (non-workshop) with the following statistics: ACL:25, EMNLP:17, NAACL:9, COLING:8, AAAI:7, NeurIPS:5, ICML:4, ICLR:3, IJCAI:3, ECIR:2, EACL:1, CVPR:1, NLPCC:1, USENIX:1, ICAART:1, Royal Society Open Science (journal):1 resulting in 89 considered papers where 60 were already surveyed by \citet{Jin:2022}.

\subsection{Human Evaluation}
\label{sec:human_evaluation_conventions}
For human evaluation, we approach the meta-analysis with the following conventions used in Tables \ref{tab:meta_analysis_1}, \ref{tab:meta_analysis_2}, and \ref{tab:meta_analysis_3}.
\begin{itemize}
    \item In the Statistical Analysis, Evaluations Released, and Adequately Specified columns, n/a refers to publications that did not conduct a human evaluation
    \item Statistical Analysis: whether either a statistical significance test on the obtained results or a power analysis in advance to estimate a sufficient number of evaluations to detect the hypothesized effect size is done
    \item Evaluations Released: whether the human evaluations have been released and are publicly available
    \item Adequately Specified: If the authors specify annotators' background, number of annotators, number of annotators per sample, incentivization, data selection, questionnaire design, rating scale
    \item Abbreviations (categorization according to \citet{Briakou:2021}) for fluency, content, style, and other: A=Absolute rating of the model output (independent from the output from other models, also includes ratings where the output sentence was rated relative to the input sentence), R=Relative to other TST models' output (e.g., ranking)
    \item Other: separate question; we do not consider combinations (e.g., geometric means) of other metrics as other
\end{itemize}

\subsection{Automated Evaluation}
\subsubsection{Conventions}
\label{sec:automated_eval_conventions}
For automated evaluation, we approach the meta-analysis with the following conventions used in Tables \ref{tab:meta_analysis_1}, \ref{tab:meta_analysis_2}, \ref{tab:meta_analysis_3}, and \ref{tab:val_overview}.
\begin{itemize}
    \item We add the automated metric to the evaluation dimension that the authors claim it measures
    \item Statistical Significance: Yes, if authors conduct a statistical significance test for their claimed results
    \item Fluency
    \begin{itemize}
        \item LM based on LSTM \cite{Hochreiter:1997} (LSTM-LM)
        \item LM based on Gated Recurrent Units \cite{Cho:2014} (GRU-LM)
        \item LM based on Recurrent Neural Networks (RNN-LM) \cite{Zhao:2018}
        \item GPT-2\cite{Radford:2019}
        \item KenLM \cite{Heafield:2011}
        \item RoBERTa \cite{Liu:2019}
        \item Statistical Grammaticality Predictor (SGP) \cite{Heilmann:2014}
        \item Kneser LM \cite{Kneser:1995}
        \item Transformer \cite{Vaswani:2017}
    \end{itemize}
    \item Content
    \begin{itemize}
        \item Cosine similarity of sentence embeddings (Cosine Sim)
        \item Cosine distance of sentence embeddings (Cosine Dist)
        \item Logistic Regression (LR)
        \item BLEU \cite{Papineni:2002}
        \begin{itemize}
            \item Source-BLEU: Comparing output to source sentences
            \item Ref-BLEU: Comparing output to reference sentences, number of references (if more than 1) in brackets
        \end{itemize}
        \item chrF \cite{Popovic:2015}
        \item BERT \cite{Devlin:2019}
        \item METEOR \cite{Banerjee:2005}
        \item BLEURT \cite{Sellam:2020}
        \item COMET \cite{Rei:2020}
        \item ROUGE \cite{Lin:2004}
        \item CIDEr \cite{Vedantam:2015}
        \item BERTScore \cite{Zhang:2020b}
        \item CNN Similarity Measure (CNN-SM) \cite{He:2015}
        \item Hu Sentiment Classifier (HSC) \cite{Hu:2016}
        \item Davidson Classifier (DC) \cite{Davidson:2017}
        \item Manning Classifier (MC) \cite{Manning:2014}
    \end{itemize}
    \item Style
    \begin{itemize}
        \item All models are classifiers if not differently specified
        \item TextCNN \cite{Kim:2014}
        \item fastText \cite{Joulin:2017}
        \item Word Mover's Distance (WMD) based on Earth Mover's Distance (EMD) \cite{Kusner:2015}
        \item Pretrained Transformer PT \cite{Wolf:2020}
        \item Formality Classifier (FC) \cite{Pavlick:2016}
    \end{itemize}
    \item Other 
    \begin{itemize}
        \item We do not consider combinations (e.g., geometric means) of other metrics as other
        \item GLEU \cite{Napoles:2015}
        \item PINC \cite{Chen:2011}
        \item Sentence Sim \cite{Wieting:2019}
        \item Flesch-Kincaid readability index \cite{Kincaid:1975}
        \item Li Diversity (LD) \cite{Li:2016b}
    \end{itemize}
\end{itemize}

\subsubsection{Ranking}
We provide an overview of the most common automated metric setups for fluency in Table~ \ref{tab:automated_fluency_ranking}, for content preservation in Table~\ref{tab:automated_content_ranking}, and for style transfer strength in Table~\ref{tab:automated_style_ranking}. 

\begin{table}[ht]
    \centering
        \begin{tabular}{p{0.35\textwidth}r}
            \textbf{Fluency} & \textbf{Count} \\
            \hline
            GPT-2 PPL & 9 \\
            GRU-LM PPL on dataset & 3\\
            5-gram KenLM PPL & 3 \\
            LSTM-LM PPL on dataset & 2\\
            RNN-LM PPL forward/backward & 2\\
            SGP & 2\\
            3-gram LM PPL on dataset & 2\\
            LSTM-LM PPL & 2 \\
            5-gram LM PPL & 2\\
            LM PPL on dataset & 2 \\
            \hline
        \end{tabular}
    \caption{Shown are the most common setups for automated fluency evaluation with at least two papers utilizing them, showing a great diversity of LM architectures and training setups (whether trained on all styles of the dataset at hand, on each style separately, forward/backward, or not further specified).}
    \label{tab:automated_fluency_ranking}
\end{table}

\begin{table}[ht]
    \centering
        \begin{tabular}{p{0.35\textwidth}r}
            \textbf{Content} & \textbf{Count} \\
            \hline
            Source-BLEU & 33 \\
            Ref-BLEU & 27\\
            Cosine Sim & 6\\
            METEOR & 4\\
            BERTScore & 4\\
            Cosine Dist & 3\\
            Word Overlap & 2\\
            BERT fine-tuned with STS & 2\\
            WMD & 2 \\
            \hline
        \end{tabular}
    \caption{Shown are the most common setups for automated content evaluation, with at least two papers utilizing them, showing the dominance of BLEU-based metrics.}
    \label{tab:automated_content_ranking}
\end{table}

\begin{table}[ht]
    \centering
        \begin{tabular}{p{0.35\textwidth}r}
            \textbf{Style} & \textbf{Count} \\
            \hline
            TextCNN & 24 \\
            fastText & 13\\
            Classifier & 10\\
            BERT & 6\\
            LSTM & 5\\
            Bi-LSTM & 3\\
            CNN & 3\\
            GRU & 3\\
            LR & 2\\
            HSC & 2\\
            RoBERTa& 2\\
            \hline
        \end{tabular}
    \caption{Shown are the most common setups for automated style transfer strength evaluation, with at least two papers utilizing them, showing the dominance of TextCNN and fastText. Often the classifier is not further specified.}
    \label{tab:automated_style_ranking}
\end{table}

\subsection{Experimentation}
\label{sec:experimentation_conventions}
For experimentation, we approach the meta-analysis with the following conventions used in Tables \ref{tab:meta_analysis_1}, \ref{tab:meta_analysis_2}, \ref{tab:meta_analysis_3}, and \ref{tab:val_overview}.
\begin{itemize}
    \item Datasets
    \begin{itemize}
        \item Amazon \cite{He:2016}
        \item Beer reviews \cite{McAuley:2012}
        \item Bible \cite{Carlson:2018}
        \item Blogs \cite{Schler:2006}
        \item Caption \cite{Li:2018}
        \item DIAL \cite{Blodgett:2016}
        \item Europarl \cite{Koehn:2005}
        \item FlickrStyle10K \cite{Gan:2017}
        \item Gender \cite{Reddy:2016}
        \item Gigaword \cite{Napoles:2012}
        \item Gutenberg \cite{Lahiri:2014}
        \item GYAFC \cite{Rao:2018}
        \item IBC \cite{Sim:2013}
        \item IMDb \cite{Diao:2014}
        \item IMDb2 \cite{Maas:2011}
        \item MSD \cite{Cao:2020}
        \item MTFC, TCFC \cite{Wu:2020}
        \item MovieTriples \cite{Serban:2016}
        \item Opinosis \cite{Ganesan:2010}
        \item Paper-News \cite{Fu:2018}
        \item Paraphrase corpus \cite{Creutz:2018}
        \item Personality captioning \cite{Shuster:2019}
        \item Political slant \cite{Voigt:2018}
        \item Reddit \cite{Santos:2018}
        \item Reddit2 \cite{Baumgartner:2020}
        \item ROC \cite{Mostafazadeh:2016}
        \item Rotten Tomatoes reviews \cite{Pang:2005}
        \item Shakespeare \cite{Xu:2012}
        \item SimpWiki \cite{Bercken:2019}
        \item SST \cite{Socher:2013}
        \item Toxicity \footnote{\url{https://www.tensorflow.org/datasets/catalog/civil_comments}}
        \item Trump speeches \footnote{\url{www.kaggle.com/binksbiz/mrtrump}}
        \item TV Series Transcripts \cite{Li:2016}
        \item Twitter \cite{Santos:2018}
        \item Twitter Persona \cite{Li:2016}
        \item Twitter Sordoni \cite{Li:2016}
        \item Twitter2 \cite{Rangel:2016}
        \item Wikipedia \cite{Xu:2016}
        \item Wikipedia2 \cite{Radford:2018}
        \item Yahoo \cite{Zhang:2015}
        \item Yelp[s] \cite{Shen:2017,Li:2018}, additional references by \citet{Jin:2019} and \citet{Luo:2019}
        \item Yelp[l] \cite{Lample:2018}, \cite{Xu:2018}, \cite{Zhang:2018}, \cite{Guu:2018}\footnote{\url{https://www.yelp.com/dataset/challenge}}
        \item Youtube2text \cite{Chen:2011}
    \end{itemize}
    \item >1 Run: No, if not mentioned or no standard deviation reported
    \item Code Provided: Yes only if code is still available
    \item Evaluation Code Provided: Yes, only if the code is still available
    \item Preprocessing Specified: No, if not explicitly mentioned
\end{itemize}


\subsection{Summary Tables}
\label{sec:summary_tables}
Table~\ref{tab:meta_analysis_1}, \ref{tab:meta_analysis_2}, and \ref{tab:meta_analysis_3} summarize our meta-analysis.

\clearpage
\begin{sidewaystable*}[htbp!]
    \centering
    \resizebox{\textwidth}{!}{
    \begin{tabular}{p{4.2cm}|p{1.8cm}p{1.8cm}p{1.8cm}p{1.2cm}p{1.2cm}p{1.2cm}p{1.8cm}|p{1.8cm}p{4cm}p{4cm}p{4cm}p{4cm}|p{6cm}p{1.5cm}p{1.5cm}p{1.5cm}p{1.5cm}}
    & \multicolumn{7}{|c|}{Human Evaluation} & \multicolumn{5}{|c|}{Automated Evaluation} & \multicolumn{5}{|c}{Experiments} \\
    \hline
    Source & \makecell[l]{Statistical\\Analysis} & \makecell[l]{Evaluations\\Released} & \makecell[l]{Adequately\\Specified} & Fluency & Content & Style & Other & \makecell[l]{Statistical\\Significance} & Fluency & Content & Style & Other & Datasets & \makecell[l]{>1 Run} & \makecell[l]{Code\\Provided} & \makecell[l]{Eval.  Code\\Provided} & \makecell[l]{Prepr.\\Specified}\\
    \hline
    \hline
        \citet{Xu:2012} & No & No & No & A & A & A & Overall A & No & - & - & Cosine sim, LM, LR & Semantic Adequacy: Ref-BLEU; Lexical Dissimilarity: PINC & Shakespeare & No & No & No & No \\
        \citet{Li:2016} & No & No & No & - & - & - & Consistency R & No & - & - & - & Overall: Ref-BLEU & Twitter Persona, Twitter Sordoni, TV Series Transcripts & No & Yes & No & No \\
        \citet{Mueller:2017} & n/a & n/a & n/a & - & - & - & - & Yes & - & - & - & Overall: LM Log Likelihood & Beer reviews, Shakespeare & Yes & No & No & No \\
        \citet{Hu:2017} & n/a & n/a & n/a & - & - & - & - & No & - & - & HSC & - & IMDb, SST & No & Yes & Yes & No \\
        \citet{Gan:2017} & No & No & No & - & - & R & - & No & - & - & - & Overall: Source-BLEU, METEOR, ROUGE, CIDEr & FlickrStyle10K, Youtube2text & No & No & No & Yes \\
        \citet{Shen:2017} & No & No & No & A & - & A & Overall R & No & - & - & TextCNN & - & Yelp[s] & No & Yes & Yes & No \\
        \citet{Han:2017} & n/a & n/a & n/a & - & - & - & - & No & - & - & - & Overall: Ref-BLEU & Shakespeare & No & No & No & No \\
        \citet{Fu:2018} & No & No & No & - & A & - & - & No & - & Cosine Dist & LSTM & - & Amazon, Paper-News & Yes & Yes & Yes & Yes \\
        \citet{Li:2018} & No & Yes & No & A & A & A & - & No & - & Ref-BLEU & Bi-LSTM & - & Yelp[s], Amazon, Captions & No & Yes & Yes & No \\
        \citet{Xu:2018} & No & Yes & No & - & A & A & - & No & - & Source-BLEU & TextCNN & - & Yelp[l], Amazon & No & Yes & Yes & No \\
        \citet{Prabhumoye:2018} & No & No & No & A & A & - & - & No & - & - & CNN & - & Gender, Political Slant, Yelp[s] & No & Yes & Yes & No \\
        \citet{Santos:2018} & n/a & n/a & n/a & - & - & - & - & No & LSTM-LM PPL on dataset & Cosine Dist & DC & - & Twitter, Reddit & No & No & No & Yes \\
        \citet{Liao:2018} & No & No & No & A & A & - & - & No & - & - & TextCNN, MC & - & Yelp[s] & No & Yes & No & No \\
        \citet{Zhang:2018} & No & No & No & A & A & A & - & No & - & Source-BLEU & TextCNN & - & Yelp[l] & No & Yes & Yes & Yes \\
        \citet{Zhao:2018} & No & No & No & A & A & A & - & No & RNN-LM PPL forward/backward & Source-BLEU & fastText & - & Yelp[s], Yahoo & No & Yes & Partially & No \\
        \citet{Rao:2018} & No & No & No & A & A & A & Overall R & No & SGP & CNN-SM & FC & Overall: Ref-BLEU, PINC, TERp & GYAFC & No & No & No & Yes \\
        \citet{Zhang:2018b} & n/a & n/a & n/a & - & - & - & - & Yes & - & - & Classifier & Overall: ROUGE & Gigaword & Yes & No & No & Yes \\
        \citet{Logeswaran:2018} & No & No & No & A & A & A & - & No & LM PPL on external dataset & Own based on BLEU, Ref-BLEU & HSC & - & Yelp[s], IMDb, Shakespeare & No & No & No & No \\
        \citet{Chen:2018} & No & No & No & A & A & A & - & No & - & - & TextCNN & Overall: Ref-BLEU; Diversity: Source-BLEU & Yelp[s] & No & Yes & Yes & Yes \\
        \citet{Yang:2018} & n/a & n/a & n/a & - & - & - & - & No & LM PPL on dataset per style & Source-BLEU & TextCNN & Overall: Ref-BLEU & Yelp[s] & No & No & No & Yes \\
        \citet{Carlson:2018} & n/a & n/a & n/a & - & - & - & - & No & - & - & - & Overall: Ref-BLEU; Diff from source:  PINC & Bible & No & Yes & Yes & Yes \\
        \citet{Niu:2018} & No & No & No & - & - & A & Overall A & No & - & Source-BLEU & SVM, CNN, LSTM-CNN & - & Stanford Politeness Corpus, MovieTriples & No & Yes & Yes & Yes \\
        \citet{Guu:2018} & No & No & No & A & A & - & Plausibility A & No & - & - & - & - & Yelp[l], Billionword & No & Yes & No & No \\
        \citet{Shetty:2018} & No & No & No & - & A & - & - & No & - & METEOR & LSTM & - & Blogs, Own & No & No & No & No \\
        \citet{Jain:2019} & No & No & No & - & - & - & Readability R & No & 4-gram KenLM PPL on  external dataset & Cosine Sim & - & Overall: Ref-BLEU; Readbility:  Flesch-Kincaid readability index & Own & No & Yes & Yes & No \\
        \citet{Dai:2019} & No & No & No & R & R & R & - & No & 5-gram KenLM PPL on dataset & Source-BLEU & fastText & Overall: Ref-BLEU & Yelp[s], IMDb & No & Yes & Yes & No \\
        \citet{John:2019} & No & No & No & A & A & A & - & No & 3-gram LM PPL on dataset & Cosine Sim, word overlap & TextCNN & - & Yelp[s], Amazon & No & Yes & Yes & No \\
        \citet{Kajiwara:2019} & n/a & n/a & n/a & - & - & - & - & No & - & - & - & Overall: Ref-BLEU, F1 on added, kept and deleted words & GYAFC & No & No & No & Yes \\
        \citet{Wu:2019} & No & No & No & A & A & A & - & No & - & Ref-BLEU & TextCNN & - & Yelp[s], Amazon & No & Yes & Yes & No \\
        \citet{Jin:2019} & No & No & No & A & A & A & - & No & LSTM-LM PPL & Ref-BLEU[5,4] & TextCNN & - & Yelp[s], GYAFC & No & No & No & No \\
        \citet{Li:2019} & No & No & No & R & R & R & Overall R & No & - & Ref-BLEU, Source-BLEU & TextCNN & - & IMDb, GYAFC, Amazon, Yelp[s], Yahoo & No & Yes & Yes & No \\
    \end{tabular}
    }
    \caption{Shown are the papers considered for the meta-analysis.}
    \label{tab:meta_analysis_1}
\end{sidewaystable*}

\clearpage
\begin{sidewaystable*}[htbp!]
    \centering
    \resizebox{\textwidth}{!}{
    \begin{tabular}{p{4.2cm}|p{1.8cm}p{1.8cm}p{1.8cm}p{1.2cm}p{1.2cm}p{1.2cm}p{1.8cm}|p{1.8cm}p{4cm}p{4cm}p{4cm}p{4cm}|p{6cm}p{1.5cm}p{1.5cm}p{1.5cm}p{1.5cm}}
    & \multicolumn{7}{|c|}{Human Evaluation} & \multicolumn{5}{|c|}{Automated Evaluation} & \multicolumn{5}{|c}{Experiments} \\
    \hline
    Source & \makecell[l]{Statistical\\Analysis} & \makecell[l]{Evaluations\\Released} & \makecell[l]{Adequately\\Specified} & Fluency & Content & Style & Other & \makecell[l]{Statistical\\Significance} & Fluency & Content & Style & Other & Datasets & \makecell[l]{>1 Run} & \makecell[l]{Code\\Provided} & \makecell[l]{Eval.  Code\\Provided} & \makecell[l]{Prepr.\\Specified}\\
    \hline
    \hline
        \citet{Shang:2019} & No & No & No & A & A & A & - & No & - & Ref-BLEU & TextCNN & Overall: GLEU & GYAFC, Own & No & No & No & Yes \\
        \citet{Sudhakar:2019} & No & No & No & R & R & - & - & No & GPT-2 PPL on dataset per style & Source-BLEU & fastText & Overall: GLEU & Yelp[s], Amazon, Captions, Political Slant, Gender & No & Yes & Yes & No \\
        \citet{Wang:2019} & n/a & n/a & n/a & - & - & - & - & No & - & BERT fine-tuned with STS & LSTM & Overall: Ref-BLEU, PINC & GYAFC & No & Yes & Partially & Yes \\
        \citet{Tikhonov:2019} & n/a & n/a & n/a & - & - & - & - & No & - & Source-BLEU & Classifier & Overall: Ref-BLEU & Yelp[s] & Yes & Yes & Yes & No \\
        \citet{Leeftink:2019} & No & No & No & - & - & A & Overall A & No & - & - & Attention-RNN & - & IMDb2, Rotten Tomatoes reviews & No & No & No & Yes \\
        \citet{Lample:2018} & No & No & No & A & A & A & Overall R & No & 5-gram Kneser LM PPL & Ref-BLEU, Source-BLEU & fastText & - & Yelp[s], Yelp[l], Amazon & No & Yes & Yes & Yes \\
        \citet{Luo:2019} & No & No & No & A & A & A & - & No & - & Ref-BLEU[4,4] & TextCNN & - & Yelp[s], GYAFC & No & Yes & Yes & No \\
        \citet{Wu:2019b} & No & No & No & A & A & A & - & No & - & Ref-BLEU & Bi-LSTM & - & Yelp[s], Amazon & No & No & No & No \\
        \citet{Fu:2019} & n/a & n/a & n/a & - & - & - & - & No & - & - & LR, TextCNN, pivot classifier & - & Yelp[s], Amazon, Captions, Political Slant, Gender, Paper-News & No & Yes & Yes & No \\
        \cite{Mir:2019} & No & Yes & No & A & A & A & - & No & LSTM-LM PPL, unigram and neural logistic regression classifiers as adversarials & Style removal, style masking with Source-BLEU, METEOR, embedding average, greedy matching, vector extrema, WMD & TextCNN, fastText & - & Yelp[s] & No & Yes & Yes & No \\
        \citet{Romanov:2019} & n/a & n/a & n/a & - & - & - & - & No & LM PPL on dataset & Cosine Sim & GRU & - & Paper-News, Shakespeare & No & Yes & No & No \\
        \citet{Gong:2019} & No & No & No & A & A & A & - & No & GRU-LM PPL on dataset per style & Cosine Dist & LSTM & - & Yelp[s], GYAFC & No & No & No & No \\
        \citet{Wang:2019b} & No & No & No & A & A & A & - & No & LM PPL on dataset & Ref-BLEU & fastText & - & Yelp[s], Amazon, Captions & No & Yes & Partially & No \\
        \citet{Li:2020} & n/a & n/a & n/a & - & - & - & - & No & - & Source-BLEU & Classifier & - & Personality captioning, FlickrStyle10K, Yahoo, Yelp[s] & No & No & No & No \\
        \citet{Liu:2020} & No & No & No & A & A & A & - & No & 3-gram Kneser LM PPL on dataset & word overlap, noun overlap & Bi-LSTM & Overall: Ref-BLEU-2 & Yelp[s], Amazon & No & Yes & Yes & Yes \\
        \citet{Pryzant:2020} & Yes & No & No & A & A & A & - & Yes & - & - & - & Overall: Source-BLEU, Classifier & Own, IBC, News headlines, Trump speeches & No & Yes & Yes & No \\
        \citet{Syed:2020} & n/a & n/a & n/a & - & - & - & - & No & - & Source-BLEU, ROUGE-1, ROUGE-2, ROUGE-3, ROUGE-L & Stylistic Alignment (Own) & - & Gutenberg, Opinosis, Wikipedia2, Shakespeare & Yes & No & No & No \\
        \citet{Wu:2020} & No & No & No & A & - & A & Overall R & No & SGP & - & GRU & Overall: Source-BLEU, BLEU-2, Embedding Avg, Embedding Extrema, Embedding Greedy & MTFC, TCFC & No & Yes & Yes & Yes \\
        \citet{Cao:2020} & No & No & No & - & A & - & - & No & BERT on dataset per style & Source-BLEU, Ref-BLEU & fastText & - & SimpWiki, MSD & No & No & No & Yes \\
        \citet{Madaan:2020} & No & No & No & A & A & A & - & No & - & Source-BLEU & LSTM & Overall: Ref-BLEU, ROUGE, METEOR & Yelp[s], Amazon, Gender, Political Slant & No & Yes & Partially & Yes \\
        \citet{Zhang:2020} & No & No & No & A & A & A & - & No & - & - & - & Overall: Ref-BLEU & GYAFC, Own & No & No & No & No \\
        \citet{Zhou:2020} & No & No & No & A & A & A & - & No & - & Ref-BLEU[4] & TextCNN & - & Yelp[s], GYAFC & No & Yes & Yes & No \\
        \citet{Jin:2020} & No & No & No & A & A & R & Attractive. A & No & GPT-2 PPL & - & - & Overall: Source-BLEU, METEOR, ROUGE, CIDEr & Own & No & Yes & Partially & No \\
        \citet{Duan:2020} & No & No & No & - & A & A & - & No & - & - & TextCNN & Diversity: LD & Yelp[s], Paper-News & No & Yes & Yes & Yes \\
        \citet{Tran:2020} & No & No & No & A & A & - & - & No & GPT-2 PPL & Source-BLEU, METEOR, ROUGE & Classifier & - & Own & No & No & No & Yes \\
        \citet{Huang:2020} & No & No & No & A & A & A & - & No & RNN-LM PPL forward/backward & Source-BLEU & fastText & - & Yelp[s], Yahoo & No & No & No & No \\
        \citet{Wang:2020} & Yes & No & No & - & - & - & Overall A & No & - & BERT fine-tuned with STS & GRU & Overall: Ref-BLEU[4], PINC & GYAFC & No & Yes & Partially & No \\
        \citet{Kim:2020} & n/a & n/a & n/a & - & - & - & - & No & 3-gram LM PPL on dataset & Source-BLEU & fastText & - & Yelp[s], Amazon & No & Yes & Yes & No \\
        \citet{Jafaritazehjani:2020} & No & No & No & A & R & A & - & No & GRU-LM PPL on dataset & Cosine Sim & TextCNN & - & Yelp[s] & No & No & No & Yes \\
        \citet{Sancheti:2020} & No & No & No & - & A & A & - & No & - & Source-BLEU & TextCNN & - & GYAFC, Yelp[l], Shakespeare & No & No & No & No \\
        \citet{Pant:2020} & n/a & n/a & n/a & - & - & - & - & No & - & Source-BLEU & Classifier & - & Yelp[s], Yelp[l] & No & No & No & No \\
        \citet{Chakrabarty:2020} & No & No & No & - & A & A & Overall A & No & - & Ref-BLEU, BERTScore & - & Novelty: Own & Own & No & Yes & Yes & Yes \\
    \end{tabular}
    }
    \caption{Shown are the papers considered for the meta-analysis.}
    \label{tab:meta_analysis_2}
\end{sidewaystable*}

\clearpage
\begin{sidewaystable*}[htbp!]
    \centering
    \resizebox{\textwidth}{!}{
    \begin{tabular}{p{4.2cm}|p{1.8cm}p{1.8cm}p{1.8cm}p{1.2cm}p{1.2cm}p{1.2cm}p{1.8cm}|p{1.8cm}p{4cm}p{4cm}p{4cm}p{4cm}|p{6cm}p{1.5cm}p{1.5cm}p{1.5cm}p{1.5cm}}
    & \multicolumn{7}{|c|}{Human Evaluation} & \multicolumn{5}{|c|}{Automated Evaluation} & \multicolumn{5}{|c}{Experiments} \\
    \hline
    Source & \makecell[l]{Statistical\\Analysis} & \makecell[l]{Evaluations\\Released} & \makecell[l]{Adequately\\Specified} & Fluency & Content & Style & Other & \makecell[l]{Statistical\\Significance} & Fluency & Content & Style & Other & Datasets & \makecell[l]{>1 Run} & \makecell[l]{Code\\Provided} & \makecell[l]{Eval.  Code\\Provided} & \makecell[l]{Prepr.\\Specified}\\
    \hline
    \hline
        \citet{Krishna:2020} & No & No & No & A & A & - & - & No & RoBERTa on external dataset & Source-BLEU & RoBERTa & - & GYAFC, Shakespeare & No & Yes & Yes & Yes \\
        \citet{Ma:2020} & No & No & No & A & R & R & - & No & Pretrained GPT PPL & BERTScore & Classifier & Repetitiveness: Own;  Diversity: Own & ROC, Paraphrase corpus & No & No & No & Yes \\
        \citet{Malmi:2020} & n/a & n/a & n/a & - & - & - & - & No & - & Source-BLEU & BERT & - & Yelp[s] & No & No & No & Yes \\
        \citet{Cheng:2020} & No & No & No & - & R & R & Context Consistency R & No & LSTM-LM PPL on dataset & Ref-BLEU & TextCNN & Overall: GLEU & GYAFC, Yahoo, Own & No & No & No & Yes \\
        \citet{Li:2020b} & n/a & n/a & n/a & - & - & - & - & No & - & Ref-BLEU, Source-BLEU & fastText & - & Yelp[s], IMDb & No & No & No & Yes \\
        \citet{Dathathri:2020} & No & No & No & A & - & R & - & No & GPT PPL & - & Classifier & Diversity: LD & n/a & No & Yes & Yes & n/a \\
        \citet{He:2020} & n/a & n/a & n/a & - & - & - & - & No & LSTM-LM PPL on dataset per style & Ref-BLEU, Source-BLEU & TextCNN & - & Yelp[s], GYAFC & No & Yes & Yes & Yes \\
        \citet{Xu:2020} & n/a & n/a & n/a & - & - & - & - & No & GPT-2 PPL & Source-BLEU & CNN & Overall: GLEU & Yelp[s], Amazon & No & Yes & Partially & Yes \\
        \citet{Yi:2020} & No & No & No & A & A & A & - & No & 5-gram KenLM PPL & Ref-BLEU, Cosine Sim & BERT & - & Yelp[s], GYAFC, Own & No & Yes & Partially & Yes \\
        \citet{Lee:2020} & No & No & No & A & A & A & - & No & GPT-1 PPL, GPT-2 PPL & Ref-BLEU[2], Source-BLEU, BERTScore & TextCNN & - & Yelp[s], Amazon & No & Yes & Yes & Yes \\
        \citet{Li:2020c} & No & No & No & A & A & - & - & No & GPT-2 PPL & Source-BLEU & RoBERTa & - & GYAFC & No & Yes & Yes & Yes \\
        \citet{Lyu:2021} & No & Yes & No & A & A & A & - & No & - & - & - & Overall: Source-BLEU, METEOR, ROUGE, CIDEr & Own & No & Yes & Partially & Yes \\
        \citet{Liu:2021} & No & No & No & A & A & A & - & No & GPT-2 PPL & Source-BLEU, Ref-BLEU & fastText & - & Yelp[s], Amazon, IMDb & No & Yes & Yes & Yes \\
        \citet{Goyal:2021} & No & No & No & A & A & A & Overall A & No & Transformer PPL on dataset & Source-BLEU, Ref-BLEU & fastText & - & GYAFC, IMDb & No & No & No & No \\
        \citet{Briakou:2021c} & No & Yes & Yes & A & A & A & Overall R & No & 5-gram Kneser LM PPL & Source-BLEU & BERT & Overall: Ref-BLEU & GYAFC, Own & No & Yes & No & Yes \\
        \citet{Ma:2021} & No & No & No & A & A & A & - & No & GRU-LM PPL on dataset & Ref-BLEU[4] & TextCNN & - & Yelp[s], GYAFC & No & Yes & Yes & No \\
        \citet{Lai:2021} & n/a & n/a & n/a & - & - & - & - & No & - & BLEURT, COMET, Ref-BLEU[4] & Classifier & - & Yelp[s], GYAFC & No & Yes & Yes & No \\
        \citet{Mireshghallah:2021} & No & No & No & R & R & - & - & No & GPT-2 PPL, GLEU, lexical diversity & Source-BLEU & Classifier & - & Yelp[s], Twitter2, DIAL & No & Yes & Yes & Yes \\
        \citet{Xiao:2021} & No & No & No & A & A & A & Consistency A & No & 5-gram KenLM PPL & Ref-BLEU, Source-BLEU & BERT & - & Yelp[s], GYAFC & No & Yes & Yes & No \\
        \citet{Ma:2021b} & No & No & No & A & A & A & - & No & GRU-LM PPL on dataset & Ref-BLEU[4] & TextCNN & - & Yelp[s], GYAFC & No & Yes & Yes & Yes \\
        \citet{Laugier:2021} & No & No & No & A & A & A & Overall A & No & GPT-2 PPL & Own & BERT & - & Yelp[s], IMDb, Toxicity & No & Yes & Yes & No \\
        \citet{Yu:2021} & No & No & No & A & A & A & - & No & Grammarly, LM PPL, adversarial unigram classifier & Attribute Hit (Own), Ref-BLEU, Source-BLEU, EMD & Classifier & - & Yelp[s], IMDb & No & No & No & Yes \\
        \citet{Briakou:2021b} & No & No & No & A & A & A & Overall R & No & LL, 5-gram KenLM PPL, BERT PLL, XLM PLL & Ref-BLEU, Source-BLEU, METEOR, chrF, WMD, Cosine Sim, BERT, BERTScore, XLM & BERT, XLM & - & GYAFC & No & Yes & Yes & No \\
        \citet{Kashyap:2022} & No & No & No & A & A & A & - & No & RoBERTa fine-tuned on external dataset & Sentence Sim & fastText & - & Yelp[s], IMDb, Political Slant & No & No & No & No \\
        \citet{Liu:2022} & No & No & No & A & A & A & - & No & - & - & TextCNN & Overall: Ref-BLEU & GYAFC & No & Yes & Yes & No \\
        \citet{Reif:2022} & No & No & No & A & A & A & - & No & GPT-2 PPL & Ref-BLEU & PT & - & Yelp[s], GYAFC & No & No & No & No \\
    \end{tabular}
    }
    \caption{Shown are the papers considered for the meta-analysis.}
    \label{tab:meta_analysis_3}
\end{sidewaystable*}

\clearpage

\section{Human Validation Details}
\label{sec:human_validation_details}

We summarize the validation of automated metrics for TST evaluation in Table~\ref{tab:val_overview}. In general, we consider an automated metric to be validated if it was validated for the mentioned aspect. We observe the three aspects of fluency, content preservation, and style transfer strength being validated. In addition, we also find validations for metrics evaluating TST as a whole. For fluency, we consider the PPL of an LM for measuring fluency to be validated. For style transfer strength, we consider any classifier architecture to be validated. Validation by \citet{Yamshchikov:2021} is not considered because of the different purpose (validated on datasets for TST and paraphrasing and not actual model outputs).

\clearpage
\begin{sidewaystable*}[t!]
   \centering
    \resizebox{\textwidth}{!}{
    \begin{tabular}{l|lllll}
    Source & Fluency & Content & Style & Overall & Datasets \\
    \hline
    \citet{Xu:2012}& - & Ref-BLEU & \makecell[l]{Ref-BLEU, PINC,\\\underline{LM}, LR, Cosine\\Sim} & - & Shakespeare \\
    \citet{Niu:2018}& - & - & - & Source-BLEU & \makecell[l]{Stanford Politeness\\Corpus}\\
    \citet{Li:2018}& Ref-BLEU & Ref-BLEU & Bi-LSTM & - & \makecell[l]{Yelp[s], Captions,\\Amazon}\\
    \citet{Fu:2018}& - & \makecell[l]{Cosine Dist} & - & - & Amazon, Paper-News\\
    \citet{Rao:2018}& SGP & CNN-SM & FC &  \underline{Ref-BLEU}, PINC, TERp & GYAFC\\
    \citet{Luo:2019}& Ref-BLEU & Ref-BLEU & TextCNN & - & Yelp[s], GYAFC\\
    \citet{Mir:2019}& \makecell[l]{\underline{Adv. Classifier},\\LSTM-LM PPL} & \makecell[l]{Source-BLEU, METEOR,\\Embed Average, Embed \\Greedy, Embed\\Extrema, \underline{WMD}} & \makecell[l]{\underline{TextCNN}, fastText} & - & Yelp[s] \\
    \citet{Pang:2019}& LM PPL & Cosine Sim & TextCNN & - & Yelp[s] \\
    \citet{Wang:2020}& - & - & - & Ref-BLEU & GYAFC\\
    \citet{Pryzant:2020}& \makecell[l]{\underline{Source-BLEU},\\Classifier} & \makecell[l]{\underline{Source-BLEU},\\Classifier} & \makecell[l]{Source-BLEU,\\\underline{Classifier}} & - & WNC \\
    \citet{Wu:2020}& SGP & - & GRU & \makecell[l]{Embed Average, Embed\\Extrema, \underline{Embed Greedy},\\BLEU-2} & MTFC, TCFC \\
    \citet{Cao:2020}& - & Ref-BLEU, Source-BLEU & - & - & SimpWiki, MSD\\
    \citet{Yamshchikov:2021}& - & \makecell[l]{Multiple} & - & - & \makecell[l]{7 TST \& Paraphrase\\Datasets}\\
    \citet{Briakou:2021b}& \makecell[l]{5-gram KenLM PPL,\\BERT PLL, \underline{XLM PLL}} & \makecell[l]{Ref-BLEU, Source-BLEU,\\METEOR, \underline{chrF}, WMD,\\Cosine Sim, BERTScore,\\BERT, XLM} & BERT, \underline{XLM} & - & GYAFC \\
    \end{tabular}
    }
    \caption{Shown are publications validating automated metrics; \underline{underlined} are the ones showing the highest correlation (if multiple compared).} 
    \label{tab:val_overview}
\end{sidewaystable*}
\clearpage

\end{document}